\documentclass[10pt,twocolumn,letterpaper]{article}
\usepackage[preprint]{cvpr}
\usepackage{amsmath,amssymb}
\usepackage{colortbl}
\usepackage{tabularx}
\newcolumntype{R}{>{\raggedleft\arraybackslash}X}
\usepackage{cuted} 
\definecolor{cvprblue}{rgb}{0.21,0.49,0.74}
\usepackage[pagebackref,breaklinks,colorlinks,allcolors=cvprblue]{hyperref}
\def\confName{CVPR}
\def\confYear{2027}
\title{LMnOP: Pruning 3D Gaussian Splatting via Leave-Many-Out Importance}
\author{
Sunjae Jeong\textsuperscript{1,2} \qquad 
Jaeseung Kim\textsuperscript{1} \qquad 
Hyoseok Hwang\textsuperscript{2, *} \vspace{0.2em} \\
\textsuperscript{1}~MOBILTECH CO., LTD. \qquad 
\textsuperscript{2}~Kyung Hee University \vspace{0.2em} \\
{\tt\small \{sunj, jason.kim\}@mobiltech.io} \qquad \tt\small hyoseok@khu.ac.kr}

\newcommand{\E}{\mathbb{E}}

\definecolor{rankgold}{HTML}{E8CE80}
\definecolor{ranksilver}{HTML}{D3DBE4}
\definecolor{rankbronze}{HTML}{E8BFA7}

\definecolor{checkgreen}{HTML}{23834B}
\newcommand{\rankfirst}[1]{\cellcolor{rankgold}\textbf{#1}}
\newcommand{\ranksecond}[1]{\cellcolor{ranksilver}\textbf{#1}}
\newcommand{\rankthird}[1]{\cellcolor{rankbronze}\textbf{#1}}
\definecolor{keptblue}{HTML}{2F7FD6}
\definecolor{remorange}{HTML}{E8862A}
\newcommand{\kbar}[1]{\textcolor{keptblue}{\rule[0.2pt]{#1\dimexpr0.42cm\relax}{3.2pt}}}
\newcommand{\rbar}[1]{\textcolor{remorange}{\rule[0.2pt]{#1\dimexpr0.42cm\relax}{3.2pt}}}
\newcommand{\yes}{\textcolor{checkgreen}{\checkmark}}
\newcommand{\pairedmethod}[1]{\raisebox{-.5\normalbaselineskip}[0pt][0pt]{#1}}
\DeclareMathSizes{5.86}{5.86}{5}{5}
\newcommand{\compacttable}{%
  \fontsize{5.86}{7.032}\selectfont\renewcommand{\arraystretch}{1}%
  \setlength{\tabcolsep}{.7pt}%
  \setlength{\heavyrulewidth}{.44pt}\setlength{\lightrulewidth}{.245pt}%
  \setlength{\cmidrulewidth}{.195pt}%
  \setlength{\aboverulesep}{1.17pt}\setlength{\belowrulesep}{1.17pt}%
  \captionsetup{skip=4pt}}

\begin{document}
\maketitle
{\renewcommand{\thefootnote}{}\footnotetext{$^{*}$Corresponding author.}}

\begin{strip}
\centering
\includegraphics[width=\linewidth]{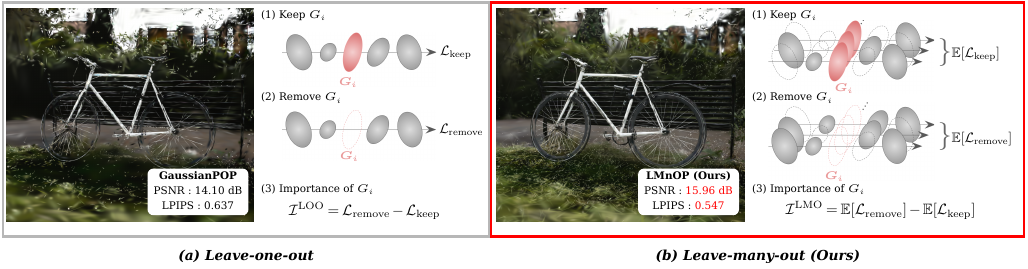}
\captionsetup{hypcap=false}
\captionof{figure}{\textbf{Pruning 99\% of the Gaussians.}
Bicycle, without fine-tuning, with PSNR and LPIPS averaged over its test views.
\textbf{(a)} The image shows the result of GaussianPOP~\cite{gaussianpop}, which uses \textbf{\textit{leave-one-out importance}}. \textit{Leave-one-out} scores each Gaussian $G_i$ by the loss increase when $G_i$ alone is removed while all the other Gaussians remain.
When several Gaussians cover the same region, each of them looks redundant on its own, so they are all removed together and leave holes.
\textbf{(b)} We therefore propose \textbf{\textit{leave-many-out}}, which scores $G_i$ by the average loss increase when the other Gaussians are also removed at random.
\textit{Leave-many-out} keeps the Gaussians that \textit{leave-one-out} would remove together.}
\label{fig:teaser}
\end{strip}

\begin{abstract}
3D Gaussian Splatting renders photorealistic views in real time.
However, a single scene often needs millions of Gaussians, so these models are costly to store and render.
Pruning reduces this cost by keeping a subset of the Gaussians.
A common score is the \textit{leave-one-out} importance, which is the loss increase when a Gaussian alone is removed.
At high pruning ratios, many Gaussians are removed together, so Gaussians that are redundant alone but essential together are lost.
We present \textbf{LMnOP} (\textbf{L}eave-\textbf{M}a\textbf{n}y-\textbf{O}ut \textbf{P}runing), which estimates the importance of each Gaussian under the removal of the other Gaussians.
This \textit{leave-many-out} importance is the average loss increase when a Gaussian is removed while the others are also removed at random.
We compute the exact mean and variance of each pixel's color under random removal in closed form, and approximate the importance from them without sampling removal patterns.
We refine each Gaussian's probability of remaining with its importance and prune once at the end.
Without fine-tuning, \textbf{LMnOP} achieves the best results in 16 of 18 metrics on three datasets at 90\% and 99\% removal.
It improves SSIM and LPIPS over GaussianPOP in all 13 scenes.
\end{abstract}

\section{Introduction}
\label{sec:intro}
 
3D Gaussian Splatting (3DGS)~\cite{kerbl2023} represents a scene with 3D Gaussians and renders photorealistic views in real time.
This combination of quality and speed has quickly made 3DGS a common choice for representing 3D scenes.
However, a single scene often requires millions of Gaussians, so the model is expensive to store and render, especially on devices with limited memory.
Several directions have been explored to reduce this cost.
One direction reduces the bits needed for each Gaussian through quantization and coding~\cite{compact3dgs,eagles,mesongs}.
This lowers storage but leaves the number of Gaussians, and thus the rendering cost, unchanged.
Another direction creates fewer Gaussians during training~\cite{minisplatting}, but this requires training the model again.
Pruning instead keeps only the Gaussians of a trained model that are actually needed.
It reduces both storage and rendering cost, and it can be combined with compression.
The more Gaussians we remove, the larger the savings, and the more it matters which Gaussians we keep.
 
To decide which Gaussians to keep, most pruning methods give each Gaussian an importance score and keep those with the highest scores.
The scores take different forms, such as accumulated blending weights~\cite{lightgaussian,minisplatting}, sensitivity to parameter changes~\cite{pup}, and the image error caused by removing a single Gaussian~\cite{gaussianpop}.
Despite their differences, each of these scores is computed while the rest of the model remains in place.
The most direct form of this idea is the loss increase when one Gaussian alone is removed.
We call this quantity its \textit{leave-one-out} importance.
In GaussianPOP~\cite{gaussianpop}, it is computed exactly for the squared error.
Such a score is accurate when only a few Gaussians are removed, because the rest of the model then barely changes.
 
However, when far more Gaussians must be removed, most of them are removed together rather than one at a time.
Consider a region that several Gaussians cover together.
If any one of them is removed, the others still fill the region, so the leave-one-out importance of each is low.
Removing all of them leaves a hole, as \cref{fig:teaser}(a) shows.
Such Gaussians are redundant alone but essential together, yet their leave-one-out importance is as low as that of truly unnecessary Gaussians.
Iterative pruning reduces this problem by removing Gaussians over several stages and scoring the remaining ones again after each stage~\cite{gaussianpop}.
 
We investigate whether the effect of removing the other Gaussians can instead be incorporated into the importance itself, before any Gaussian is removed.
A Gaussian should be evaluated in the situation that pruning actually creates, where many of the other Gaussians are removed as well.
Following this principle, we present \textit{LMnOP}, which scores each Gaussian by its \textit{leave-many-out} (LMO) importance (\cref{fig:teaser}(b)).
This importance is the average loss increase when the Gaussian is removed while the other Gaussians are also removed at random, each according to its keep probability.
When every keep probability is one, LMO importance reduces to leave-one-out importance.
The keep probabilities should reflect which Gaussians will remain, but this is not known in advance.
We therefore refine them with the importance over several steps, so that each Gaussian is evaluated in a situation close to the final selection.
No Gaussian is removed during these steps, so the selection can still change.
One final pruning step then keeps the Gaussians with the highest importance.
 
Computing LMO importance is the main difficulty.
It is an average over all removal patterns, and rendering each pattern would be prohibitively expensive.
We introduce $\mu$-$\sigma$ rendering, which computes the exact mean and variance of each pixel's color under random removal in closed form.
From these two statistics, we compute the importance of all Gaussians without sampling removal patterns.
The importance is exact for the squared error and approximate for the loss that we use.
 
Our contributions are as follows:
\begin{itemize}
\item We define LMO importance to evaluate each Gaussian while the other Gaussians are also removed at random. We refine the keep probabilities with this importance and prune once at the end.
\item We introduce $\mu$-$\sigma$ rendering, which computes the exact mean and variance of each pixel's color under random removal in closed form. With these exact statistics, we approximate the importance of all Gaussians without sampling removal patterns.
\item Without fine-tuning, \textit{LMnOP} achieves the best results in 16 of 18 metrics across three datasets at 90\% and 99\% removal. We also show that it keeps Gaussians whose importance rises after pruning.
\end{itemize}
 
\section{Related work}
\label{sec:related}
 
\begin{figure*}[t]
\centering
\includegraphics[width=\linewidth]{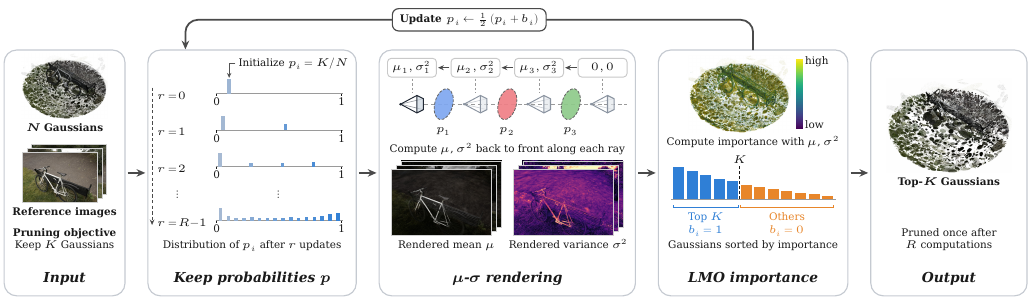}
\caption{\textbf{Overview of \textit{LMnOP}.}
Every Gaussian starts with the keep probability $K/N$, and the histograms show the keep probabilities after $r$ updates.
$\mu$-$\sigma$ rendering gives the mean and variance of each pixel's color under the current keep probabilities, and we compute the LMO importance of every Gaussian from them.
We then move each keep probability halfway toward $b_i$ and compute the importance again.
Here $b_i$ is $1$ for the $K$ Gaussians with the highest importance and $0$ for the others.
After $R$ importance computations, we keep these $K$ Gaussians and prune the rest in one final step.}
\label{fig:overview}
\end{figure*}
 
\paragraph{Compact 3DGS.}
3DGS models~\cite{kerbl2023} can be made compact in two complementary ways.
One reduces the bits needed for each Gaussian, and the other reduces the number of Gaussians.
Compact3DGS~\cite{compact3dgs}, EAGLES~\cite{eagles}, and MesonGS~\cite{mesongs} combine compact attribute representations with mechanisms that reduce model size.
Mini-Splatting~\cite{minisplatting} instead creates fewer Gaussians during training.
Pruning a trained model is the most direct way to reduce the number of Gaussians, and it can be combined with attribute compression.
Its result, however, depends entirely on which Gaussians are judged important.
 
\paragraph{Importance-based pruning.}
Most pruning methods assign a score to each Gaussian and keep the highest-scoring ones.
LightGaussian~\cite{lightgaussian} and Mini-Splatting measure how much each Gaussian contributes to rendering.
PUP 3D-GS~\cite{pup} approximates the sensitivity of the loss to second order.
GaussianPOP~\cite{gaussianpop} analytically computes the error caused by removing a single Gaussian.
Trimming~\cite{trim} uses gradient information, and RAP~\cite{rap} predicts importance from Gaussian attributes and neighborhood statistics.
Other methods learn masks instead of computing scores~\cite{compact3dgs,maskgaussian}, but they must optimize the model again.
MaskGaussian, for example, samples probabilistic masks during optimization and keeps gradients for masked Gaussians.
The scores above differ in form, but each of them is computed while the rest of the model is present.
As a result, they reflect what a Gaussian contributes alone, not what happens when many Gaussians are removed together.
 
\paragraph{Interaction-aware pruning.}
What happens when Gaussians are removed together depends on how they interact, since removing one changes the transmittance of every Gaussian behind it.
The same issue arises in network pruning, where Optimal Brain Surgeon~\cite{obs} and its combinatorial extension~\cite{cbs} account for interactions between weights.
Within 3DGS, prior work handles the interaction in three ways.
REFINE~\cite{refine} approximates the importance without rendering, under the assumption that the Gaussians are independent.
The rendering guarantee of coresets~\cite{coreset} relies on assumptions of validity and log-transmittance stability.
GaussianPOP repeats scoring over multiple pruning stages, so that later scores reflect earlier removals.
Concurrent work, SplatQ~\cite{splatq}, models the effect of removing pairs of Gaussians within spatial blocks and solves a quadratic selection problem.
The interaction is thus assumed away, captured only across pruning stages, or modeled only between pairs within blocks.
We instead score each Gaussian by its LMO importance.
It averages over the random removal of all the other Gaussians.
The interactions among all Gaussians on each ray are thus taken into account in the importance itself, before any Gaussian is removed.
 
\section{Preliminaries}
\label{sec:prelim}
 
\subsection{3DGS rendering}
\label{sec:prelim_3dgs}
3DGS~\cite{kerbl2023} represents a scene with $N$ Gaussians, each with a position, a covariance, an opacity, and a view-dependent color.
To render a pixel, the Gaussians that cover it are sorted from front to back and indexed by $k=1,\ldots,n$.
The $k$-th Gaussian has color $c_k$ and alpha $\alpha_k$, which is its opacity multiplied by its projected 2D Gaussian at the pixel.
The pixel color is then
\begin{equation}
 C=\sum_{k=1}^{n}\alpha_k c_k T_k,\quad\text{with}\quad T_k=\prod_{j<k}(1-\alpha_j),
 \label{eq:render}
\end{equation}
where the transmittance $T_k$ is the fraction of light that reaches the $k$-th Gaussian.
Since a Gaussian appears at different depths on different rays, we write $i$ for a Gaussian in the model and $k$ for its position on the current ray.
We write all equations for one color channel and assume a black background.
 
\subsection{Pruning objective}
\label{sec:prelim_objective}
Pruning keeps $K$ of the $N$ Gaussians without changing their attributes.
We describe which Gaussians are kept with a binary mask $m=(m_1,\ldots,m_N)$, where $m_i=1$ keeps Gaussian $i$ and $m_i=0$ removes it.
In \cref{eq:render}, this amounts to replacing each $\alpha_k$ with $m_k\alpha_k$, so a removed Gaussian contributes nothing.
To measure the damage of pruning, we compare the images rendered with $m$ to reference images.
A reference image is the render of the original model from a training view, and $C^{\mathrm{ref}}$ is its color at a pixel.
The reference images stay fixed throughout pruning.
The pruning loss $\mathcal L(m)$ averages this comparison over all pixels and views, using an L1 term and a D-SSIM term.
The pruning problem is
\begin{equation}
 \min_{m\in\{0,1\}^N}\;\mathcal L(m)\quad\text{with}\quad\textstyle\sum_{i}m_i=K.
 \label{eq:objective}
\end{equation}
There are far too many possible masks to search directly, so most methods instead score each Gaussian and keep the $K$ highest-scoring ones.
A common score is the \textit{leave-one-out} importance, which is the loss increase when only Gaussian $i$ is removed from the original model.
 
\section{Method}
\label{sec:method}
 
\subsection{Leave-many-out importance}
\label{sec:lmo}
Leave-one-out importance is measured while all the other Gaussians remain.
We instead evaluate each Gaussian while the other Gaussians are also removed at random.
We treat each $m_i$ as an independent random variable that equals $1$ with probability $p_i$.
We call $p_i$ the keep probability of Gaussian $i$.
The \textit{leave-many-out} (LMO) importance of Gaussian $i$ is
\begin{equation}
 \mathcal I_i(p)=\E[\mathcal L\mid m_i=0]-\E[\mathcal L\mid m_i=1],
 \label{eq:lmo}
\end{equation}
where the expectation is over the other Gaussians.
A large value means that keeping Gaussian $i$ reduces the expected loss.
When all the other Gaussians remain, the LMO importance equals the leave-one-out importance.
The simplest choice of keep probabilities is $K/N$ for every Gaussian, but important Gaussians are more likely to remain than the others.

\subsection{Importance-aware keep probabilities}
\label{sec:refine}
Important Gaussians should receive high keep probabilities, but the importance itself depends on the keep probabilities.
We resolve this circular dependence with an update that is motivated by the expected loss when $K$ Gaussians remain on average.
The LMO importance is the negative derivative of the expected loss with respect to $p_i$.
To first order, this expected loss is therefore lowest when the keep probabilities are $1$ for the $K$ Gaussians with the highest importance and $0$ for the others.
Let $b_i$ denote this value for Gaussian $i$.
This estimate changes as the keep probabilities change, so we move only halfway toward it:
\begin{equation}
 p_i\leftarrow\tfrac12\,(p_i+b_i).
 \label{eq:update}
\end{equation}
The update keeps the sum of the keep probabilities at $K$.
Starting from $K/N$ for every Gaussian, we alternate between computing the importance and applying \cref{eq:update}.
We compute the importance $R$ times in total, with $R-1$ updates in between (\cref{fig:overview}).
With each update, the keep probabilities move from $K/N$ toward $0$ or $1$.
No Gaussian is removed during these steps, so the selection can still change.
After the last computation, we keep the $K$ Gaussians with the highest LMO importance and prune the rest in one final step.
The random masks serve only to compute the importance.

\subsection{\texorpdfstring{$\mu$-$\sigma$ rendering}{mu-sigma rendering}}
\label{sec:musigma}
It remains to compute the LMO importance.
Evaluating the expectation in \cref{eq:lmo} directly would require rendering every possible mask.
For the squared error, however, the expected loss at a pixel depends only on the mean $\mu$ and variance $\sigma^2$ of its color:
\begin{equation}
 \E\big[(C-C^{\mathrm{ref}})^2\big]=(\mu-C^{\mathrm{ref}})^2+\sigma^2.
 \label{eq:decomp}
\end{equation}
We compute both statistics from back to front along each ray, as \cref{fig:overview} illustrates.
Let $\mu_k$ and $\sigma_k^2$ be the mean and variance of the color blended from the $k$-th Gaussian to the last one.
Keeping the $k$-th Gaussian instead of removing it changes the mean by $\delta_k=\alpha_k(c_k-\mu_{k+1})$.
We write $\gamma_k = 1 - p_k\alpha_k(2-\alpha_k)$ for the expected squared transmittance through this Gaussian.
The two statistics then follow
\begin{equation}
 \begin{aligned}
 \mu_k&=\mu_{k+1}+p_k\delta_k,\\
 \sigma_k^2&=\gamma_k\sigma_{k+1}^2+p_k(1-p_k)\delta_k^2.
 \end{aligned}
 \label{eq:recursion}
\end{equation}
Starting from zero behind the last Gaussian, the recursion ends with the mean and variance of the pixel, which are exact for independent masks.
We call this computation $\mu$-$\sigma$ rendering.
Its mean is $\alpha$-blending with each alpha scaled by its keep probability.
Its variance measures how much the pixel changes when Gaussians are removed together.
 
\subsection{Scoring loss}
\label{sec:objective}
The mean and variance give the expected loss exactly only for the squared error.
In practice, we follow the 3DGS training loss~\cite{kerbl2023} and combine an L1 term and a D-SSIM term with weights $1-\lambda$ and $\lambda$.
We call this combination the scoring loss.
For the L1 term, we measure each pixel with
\begin{equation}
 e=\sqrt{(\mu-C^{\mathrm{ref}})^2+\beta\sigma^2+\varepsilon},
 \label{eq:loss}
\end{equation}
where $\varepsilon$ is a small constant, so $e$ is a smoothed absolute error when the variance is zero.
We average $e$ over color channels, pixels, and views to obtain $\mathcal L_1$, and $\beta$ sets how much the variance counts.
The D-SSIM term $\mathcal L_{\text{D-SSIM}}$ is computed between the mean image and the reference image.
When every keep probability is 0 or 1, the variance vanishes and the scoring loss equals the pruning loss in \cref{eq:objective}, up to the smoothing by $\varepsilon$.
For other keep probabilities, it is a surrogate for the expected loss.
 
We compute the LMO importance of the scoring loss as
\begin{equation}
 \mathcal I_i(p)\approx(1-\lambda)\,\Delta_i\mathcal L_1+\lambda\,\Delta_i\mathcal L_{\text{D-SSIM}},
 \label{eq:importance}
\end{equation}
where $\Delta_i$ is the change in a loss when Gaussian $i$ is removed instead of kept, with the other keep probabilities fixed.
The pixel statistics with Gaussian $i$ kept or removed follow in closed form from one front-to-back and one back-to-front pass, without rendering each Gaussian separately.
This gives $\Delta_i\mathcal L_1$ exactly.
For the D-SSIM term, we expand the loss to second order around the mean image, and the second-order term is weighted by $p_i-\tfrac12$.
We use this approximate importance to compute the update in \cref{eq:update}.

\begin{table*}[!t]
\centering\compacttable
\caption{\textbf{Comparison with baselines.}
Three datasets at 90\% and 99\% removal, without fine-tuning.
All methods prune the same original model to the same number of Gaussians.
Throughout the paper, gold, silver, and bronze mark the three best values of a column before rounding, and tied values share a rank.}
\label{tab:main}
\begin{tabularx}{\linewidth}{@{}l*{18}{R}}
\toprule
 & \multicolumn{9}{c}{90\% removed} & \multicolumn{9}{c}{99\% removed} \\
\cmidrule(lr){2-10}\cmidrule(lr){11-19}
 & \multicolumn{3}{c}{Mip-NeRF360} & \multicolumn{3}{c}{Deep Blending} & \multicolumn{3}{c}{Tanks \& Temples} & \multicolumn{3}{c}{Mip-NeRF360} & \multicolumn{3}{c}{Deep Blending} & \multicolumn{3}{c}{Tanks \& Temples} \\
\cmidrule(lr){2-4}\cmidrule(lr){5-7}\cmidrule(lr){8-10}\cmidrule(lr){11-13}\cmidrule(lr){14-16}\cmidrule(lr){17-19}
Method & PSNR$\uparrow$ & SSIM$\uparrow$ & LPIPS$\downarrow$ & PSNR$\uparrow$ & SSIM$\uparrow$ & LPIPS$\downarrow$ & PSNR$\uparrow$ & SSIM$\uparrow$ & LPIPS$\downarrow$ & PSNR$\uparrow$ & SSIM$\uparrow$ & LPIPS$\downarrow$ & PSNR$\uparrow$ & SSIM$\uparrow$ & LPIPS$\downarrow$ & PSNR$\uparrow$ & SSIM$\uparrow$ & LPIPS$\downarrow$ \\
\midrule
Original (0\%) & 27.35 & 0.814 & 0.257 & 29.52 & 0.903 & 0.311 & 23.39 & 0.842 & 0.219 & 27.35 & 0.814 & 0.257 & 29.52 & 0.903 & 0.311 & 23.39 & 0.842 & 0.219 \\
\midrule
Uniform & 12.52 & 0.271 & 0.612 & 12.61 & 0.445 & 0.628 & 9.91 & 0.279 & 0.599 & 8.87 & 0.061 & 0.716 & 7.48 & 0.104 & 0.752 & 5.77 & 0.057 & 0.708 \\
Opacity & 15.36 & 0.378 & 0.561 & 17.91 & 0.621 & 0.539 & 13.91 & 0.452 & 0.512 & 9.76 & 0.125 & 0.701 & 7.93 & 0.135 & 0.745 & 7.00 & 0.141 & 0.691 \\
EAGLES$^{*}$~\cite{eagles} & \rankthird{20.29} & \rankthird{0.651} & \ranksecond{0.394} & 25.04 & \rankthird{0.835} & \rankthird{0.400} & 17.88 & \rankthird{0.689} & \rankthird{0.355} & 14.20 & \ranksecond{0.413} & \ranksecond{0.621} & 16.61 & 0.692 & 0.577 & 12.30 & \ranksecond{0.458} & \rankthird{0.610} \\
MesonGS$^{*}$~\cite{mesongs} & 19.86 & 0.630 & \rankthird{0.408} & 24.88 & 0.832 & 0.404 & 18.07 & 0.678 & 0.368 & 14.26 & \rankthird{0.409} & 0.634 & 16.96 & \rankthird{0.700} & 0.580 & \rankthird{12.47} & \rankthird{0.455} & 0.619 \\
Trimming$^{*}$~\cite{trim} & 15.55 & 0.422 & 0.532 & 17.68 & 0.655 & 0.509 & 14.08 & 0.484 & 0.483 & 9.67 & 0.134 & 0.696 & 8.00 & 0.149 & 0.720 & 6.82 & 0.145 & 0.674 \\
LightGaussian$^{*}$~\cite{lightgaussian} & 19.16 & 0.578 & 0.450 & 24.34 & 0.817 & 0.425 & 17.66 & 0.649 & 0.394 & \rankthird{14.33} & 0.393 & 0.646 & \rankthird{18.22} & \ranksecond{0.713} & 0.584 & 12.41 & 0.447 & 0.622 \\
PUP 3D-GS$^{*}$~\cite{pup} & 17.05 & 0.578 & 0.442 & 22.55 & 0.798 & 0.432 & 13.55 & 0.601 & 0.414 & 11.18 & 0.256 & 0.653 & 11.31 & 0.479 & 0.643 & 8.91 & 0.320 & 0.631 \\
GaussianPOP$^{\dagger}$~\cite{gaussianpop} & \ranksecond{22.55} & \ranksecond{0.664} & 0.409 & \rankfirst{27.69} & \ranksecond{0.854} & \ranksecond{0.388} & \rankfirst{20.81} & \ranksecond{0.718} & \ranksecond{0.355} & \ranksecond{15.38} & 0.376 & \rankthird{0.628} & \ranksecond{19.86} & 0.687 & \ranksecond{0.563} & \ranksecond{14.00} & 0.443 & \ranksecond{0.586} \\
RAP~\cite{rap} & 19.89 & 0.568 & 0.468 & \rankthird{25.40} & 0.823 & 0.432 & \rankthird{18.33} & 0.641 & 0.411 & 13.88 & 0.347 & 0.660 & 18.13 & 0.690 & \rankthird{0.576} & 12.24 & 0.408 & 0.630 \\
Coreset$^{*}$~\cite{coreset} & 20.22 & 0.646 & 0.412 & 23.89 & 0.821 & 0.419 & 17.65 & 0.658 & 0.389 & 14.32 & 0.403 & 0.634 & 16.88 & 0.681 & 0.579 & 12.36 & 0.444 & 0.619 \\
\midrule
\textbf{LMnOP (Ours)} & \rankfirst{23.00} & \rankfirst{0.715} & \rankfirst{0.373} & \ranksecond{27.22} & \rankfirst{0.869} & \rankfirst{0.371} & \ranksecond{20.15} & \rankfirst{0.736} & \rankfirst{0.332} & \rankfirst{16.53} & \rankfirst{0.467} & \rankfirst{0.594} & \rankfirst{19.97} & \rankfirst{0.732} & \rankfirst{0.551} & \rankfirst{14.63} & \rankfirst{0.512} & \rankfirst{0.550} \\
\bottomrule
\end{tabularx}
\vspace{1pt}
{\raggedleft\scriptsize $^{\dagger}$Our implementation.\quad $^{*}$Importance or selection component only.\par}
\end{table*}

\section{Experiments}
\label{sec:experiments}

\subsection{Setup}
\label{sec:setup}

\paragraph{Datasets and metrics.}
We evaluated 13 scenes: 9 from Mip-NeRF360~\cite{mipnerf360}, 2 from Deep Blending~\cite{deepblending}, and 2 from Tanks \& Temples~\cite{tanks}.
For each scene, the original model was a 3DGS model~\cite{kerbl2023} trained at the standard dataset resolution.
We measured PSNR, SSIM~\cite{ssim}, and LPIPS~\cite{lpips} on the test views and averaged them over the scenes of each dataset with equal weight.
We computed LPIPS with the VGG network on images scaled to $[-1,1]$.

\paragraph{Baselines.}
All methods pruned the same original model to the same number of Gaussians, and no method was fine-tuned afterward.
Several baselines are parts of larger pipelines that also retrain or compress the model.
For these methods, we applied only their importance or selection component.
Uniform keeps $K$ Gaussians chosen at random, and Opacity keeps the $K$ most opaque ones.
For GaussianPOP~\cite{gaussianpop}, we used our own implementation with eight pruning stages, as in its paper.

\paragraph{Implementation details.}
\textit{LMnOP} computed the importance $R=8$ times while keeping all the Gaussians and then pruned each model in one final step.
We set $\lambda=0.2$, $\beta=0.5$, and $\varepsilon=10^{-6}$.
The importance was accumulated over all training views, with the renders of the original model as reference images.
These settings were the same for all scenes and pruning ratios.
We ran all experiments on an NVIDIA RTX A6000 GPU.

\subsection{Comparison with baselines}
\label{sec:comparison}
We compared \textit{LMnOP} with the baselines at 90\% and 99\% removal.
No model was fine-tuned after pruning, so the results reflect only which Gaussians each method kept.

\paragraph{Quantitative results.}
As \cref{tab:main} shows, \textit{LMnOP} achieved the best value in 16 of the 18 columns.
GaussianPOP and EAGLES~\cite{eagles} followed in most columns.
The two exceptions were the PSNR at 90\% removal on two datasets, where GaussianPOP was higher.
In SSIM and LPIPS, \textit{LMnOP} was better than GaussianPOP in all 13 scenes at both pruning ratios.
At 99\% removal, it was the best of all methods in both metrics in every scene.

\begin{figure}[!t]
\centering\includegraphics[width=\linewidth]{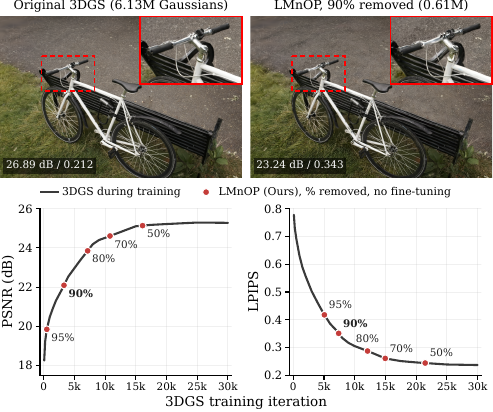}
\caption{\textbf{Quality of pruned models on the 3DGS training curve.}
Bicycle, without fine-tuning.
\textbf{Top}: the original model and \textit{LMnOP} at 90\% removal, with the PSNR\,/\,LPIPS of the view.
The zoomed region in the red box shows slight white artifacts in the pruned model.
\textbf{Bottom}: test PSNR and LPIPS of 3DGS during training.
Each marker places a pruned model where 3DGS reaches the same value.
The markers indicate quality, not training time.}
\label{fig:training_reference}
\end{figure}

The metrics alone do not show how close a pruned model is to the original model.
We therefore compared a pruned model with the training curve of 3DGS (\cref{fig:training_reference}).
On Bicycle, the model with 90\% of the Gaussians removed matched the LPIPS that 3DGS reaches after about 7.4k of its 30k training iterations.
It matched the PSNR that 3DGS reaches after about 3.3k iterations.
In the example view, its render kept the overall appearance of the original, although it kept only a tenth of the Gaussians and was not fine-tuned.

\begin{figure*}[!t]
\centering
\begin{subfigure}{\linewidth}
\centering\includegraphics[width=\linewidth]{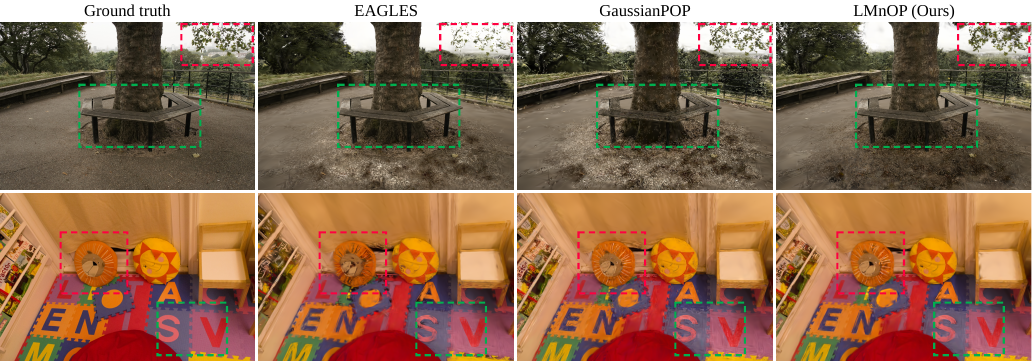}
\caption{90\% removed}
\label{fig:qualitative_90}
\end{subfigure}
\par\smallskip
\begin{subfigure}{\linewidth}
\centering\includegraphics[width=\linewidth]{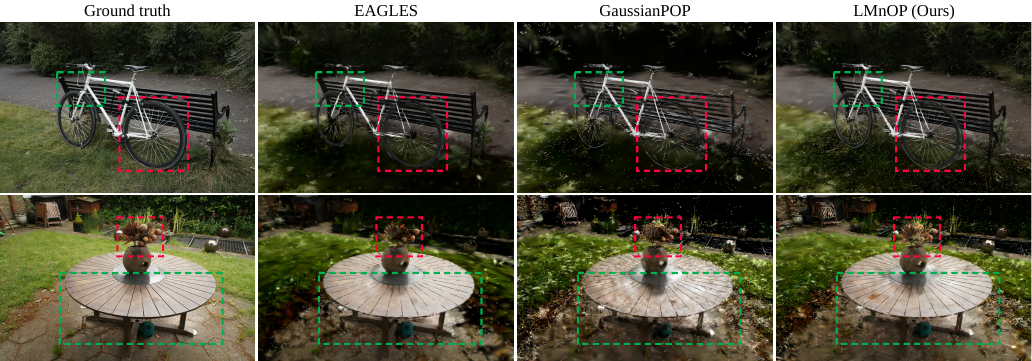}
\caption{99\% removed}
\label{fig:qualitative_99}
\end{subfigure}
\caption{\textbf{Qualitative comparison.}
(a) Treehill and Playroom at 90\% removal and (b) Bicycle and Garden at 99\% removal, without fine-tuning.
Red boxes mark high-frequency regions, and green boxes mark low-frequency regions.}
\label{fig:qualitative}
\end{figure*}

\paragraph{Qualitative results.}
In \cref{fig:qualitative}, we compare \textit{LMnOP} with EAGLES and GaussianPOP, the two strongest baselines in \cref{tab:main}.
Red boxes mark high-frequency regions with fine detail, and green boxes mark low-frequency regions with smooth surfaces.
The two baselines failed in opposite regions.
For example, on Treehill in the top row of \cref{fig:qualitative}(a), EAGLES rendered the bench well but not the leaves.
GaussianPOP rendered the leaves well but produced artifacts on the bench.
The other scenes showed the same pattern.
At 99\% removal, GaussianPOP did not render the table of Garden well (bottom row of \cref{fig:qualitative}(b)).
\textit{LMnOP} rendered both the high-frequency and the low-frequency regions well in all four scenes.

\subsection{Importance before and after pruning}
\label{sec:before_after}
The comparison above shows that \textit{LMnOP} selects better subsets, but it does not show why.
Our premise is that the importance of a Gaussian changes once many other Gaussians are removed.
Leave-one-out importance is measured in the original model, so it does not reflect such a change.
We tested this premise directly, with Gaussians that have the same leave-one-out importance.

\paragraph{Design.}
We pruned the original model with leave-one-out importance under our scoring loss, in a single stage, and considered the Gaussians that were removed. We call the result the pruned model.
\textit{LMnOP} kept some of these Gaussians and removed the others.
We formed pairs with the same leave-one-out importance, each with one Gaussian that \textit{LMnOP} kept and one that it removed.
The two Gaussians of a pair were therefore equally unimportant before pruning and differed only in whether \textit{LMnOP} kept them.
If pruning did not change the importance, they would remain equally important afterward.
We measured the importance of both Gaussians in two models, by rendering, with the same loss and reference images.
In the original model, we removed the Gaussian alone and recorded the loss increase.
In the pruned model, we restored the Gaussian alone and recorded the loss decrease.
We measured 500 pairs per condition, chosen before any measurement.

\begin{table}[!t]
\centering\compacttable
\caption{\textbf{Importance of paired Gaussians in the original and the pruned model.}
Each pair has the same leave-one-out importance.
\textit{LMnOP} keeps one Gaussian of each pair and removes the other.
Values are medians ($\times10^{-6}$). Bars show each pair on a common scale.}
\label{tab:before_after}
\begin{tabular*}{\linewidth}{@{\extracolsep{\fill}}llr@{\,}lr@{\,}lr@{\,}lr@{\,}lc@{}}
\toprule
 & & \multicolumn{4}{c}{Original model} & \multicolumn{5}{c}{Pruned model} \\
\cmidrule(lr){3-6}\cmidrule(lr){7-11}
Removal & Scene & \multicolumn{2}{c}{Kept} & \multicolumn{2}{c}{Removed} & \multicolumn{2}{c}{Kept} & \multicolumn{2}{c}{Removed} & Ratio \\
\midrule
90\% & Bicycle & 0.035 & \kbar{1.00} & 0.033 & \rbar{0.94} & \textbf{0.167} & \kbar{1.00} & 0.055 & \rbar{0.33} & \textbf{3.0$\times$} \\
\midrule
99\% & Bicycle & 0.104 & \kbar{1.00} & 0.102 & \rbar{0.98} & \textbf{1.295} & \kbar{1.00} & 0.191 & \rbar{0.15} & \textbf{6.8$\times$} \\
 & Kitchen & 0.209 & \kbar{1.00} & 0.203 & \rbar{0.97} & \textbf{4.285} & \kbar{1.00} & 0.805 & \rbar{0.19} & \textbf{5.3$\times$} \\
 & Garden & 0.105 & \kbar{1.00} & 0.100 & \rbar{0.95} & \textbf{1.470} & \kbar{1.00} & 0.291 & \rbar{0.20} & \textbf{5.0$\times$} \\
\bottomrule
\end{tabular*}
\end{table}

\paragraph{Results.}
In the original model, the two Gaussians of a pair were nearly equally important, as intended (\cref{tab:before_after}).
In the pruned model, they were not.
The Gaussians that \textit{LMnOP} kept were 3.0 times as important as the ones that it removed after 90\% removal and 5.0 to 6.8 times after 99\%.
They were the more important ones in 88--94\% of the pairs.
Restoring all the kept Gaussians to the pruned model raised PSNR by 1.7--3.3\,dB, whereas restoring the same number of removed ones raised it by 0.4--0.8\,dB.
One may ask whether \textit{LMnOP} simply kept larger or more opaque Gaussians.
The difference remained when the pairs were also matched in opacity, size, and number of visible views.

The importance of a Gaussian therefore depends on which Gaussians remain.
Leave-one-out importance is measured before pruning and does not reflect this change, whereas \textit{LMnOP} keeps the Gaussians that become important.
This test concerns a leave-one-out ranking that is computed once before pruning.

\subsection{Robustness to the original model}
\label{sec:base_models}
So far, the original models were trained with 3DGS.
LMO importance differs from leave-one-out importance when several Gaussians cover the same region.
How the Gaussians cover a scene depends on how the original model was trained.
We therefore tested whether the advantage of \textit{LMnOP} holds for original models trained in different ways.
We pruned original models trained with six methods on the Mip-NeRF360 scenes and compared \textit{LMnOP} with five baselines (\cref{tab:bases}).

\paragraph{Training methods.}
3DGS~\cite{kerbl2023} is the reference and grows Gaussians by cloning and splitting.
The other five methods each change one of three properties.
The first is where the Gaussians are placed.
MCMC~\cite{mcmc} places them by sampling and relocation.
We set its number of Gaussians to that of 3DGS in each scene.
The two models thus differ in where the Gaussians are, not in how many there are.
The second is how the Gaussians are grown.
AbsGS~\cite{absgs} and Pixel-GS~\cite{pixelgs} change the rule for where new Gaussians are added.
Mini-Splatting-D~\cite{minisplatting} is the dense model of Mini-Splatting before its simplification.
These three methods are designed for finer detail and better spatial coverage.
The third is how compact the model already is.
FastGS~\cite{fastgs} prunes Gaussians during training, so its model has far fewer Gaussians than the others.

\paragraph{Results.}
\textit{LMnOP} achieved the lowest LPIPS in 11 of the 12 conditions.
On MCMC at 99\% removal, it lowered LPIPS from 0.620, the best baseline, to 0.590.
It was also the best at both pruning ratios on AbsGS, Pixel-GS, and Mini-Splatting-D.
The advantage of \textit{LMnOP} therefore does not depend on where the Gaussians were placed or how they were grown.

The one exception was FastGS at 99\% removal, where GaussianPOP and \textit{LMnOP} were nearly tied at 0.712 and 0.714.
This condition is far more extreme than the others.
Because FastGS is already compact, 99\% removal left only about 4{,}000 Gaussians per scene, compared with about 34{,}000 for 3DGS.
With so few Gaussians, all methods had an LPIPS above 0.71 and differed by at most 0.020.
We attribute the missing advantage to the compactness of FastGS.
LMO importance differs from leave-one-out importance when several Gaussians cover the same region.
A model that was already pruned during training has few such Gaussians left.
At 90\% removal on the same model, \textit{LMnOP} was again the best, with 0.535 against 0.561.

\begin{table}[!t]
\centering\compacttable
\caption{\textbf{Pruning models trained with different methods.}
LPIPS ($\downarrow$) on Mip-NeRF360 at 90\% and 99\% removal, without fine-tuning.
The first row is the original model.}
\label{tab:bases}
\begin{tabularx}{\linewidth}{@{}lc*{6}{R}}
\toprule
 &  & \multicolumn{6}{c}{Training method} \\
\cmidrule(lr){3-8}
Method & Rem. & 3DGS & MCMC & FastGS & \shortstack{Mini-\\Splatting-D} & AbsGS & Pixel-GS \\
\midrule
Original & 0\% & 0.257 & 0.224 & 0.306 & 0.210 & 0.229 & 0.227 \\
\midrule
\pairedmethod{EAGLES$^{*}$} & 90\% & \ranksecond{0.394} & \rankthird{0.428} & \rankthird{0.568} & \rankthird{0.428} & \ranksecond{0.420} & \ranksecond{0.369} \\
 & 99\% & \ranksecond{0.621} & \rankthird{0.636} & \rankthird{0.721} & \rankthird{0.645} & \ranksecond{0.647} & \ranksecond{0.606} \\
\pairedmethod{MesonGS$^{*}$} & 90\% & \rankthird{0.408} & 0.433 & 0.583 & 0.453 & 0.441 & 0.382 \\
 & 99\% & 0.634 & 0.638 & 0.730 & 0.665 & 0.658 & 0.622 \\
\pairedmethod{GaussianPOP$^{\dagger}$} & 90\% & 0.409 & \ranksecond{0.410} & \ranksecond{0.561} & \ranksecond{0.420} & 0.455 & \rankthird{0.382} \\
 & 99\% & \rankthird{0.628} & \ranksecond{0.620} & \rankfirst{0.712} & \ranksecond{0.643} & \rankthird{0.653} & \rankthird{0.612} \\
\pairedmethod{RAP} & 90\% & 0.468 & 0.469 & 0.583 & 0.472 & 0.516 & 0.428 \\
 & 99\% & 0.660 & 0.658 & 0.732 & 0.663 & 0.691 & 0.638 \\
\pairedmethod{Coreset$^{*}$} & 90\% & 0.412 & 0.440 & 0.578 & 0.468 & \rankthird{0.439} & 0.391 \\
 & 99\% & 0.634 & 0.650 & 0.729 & 0.667 & 0.661 & 0.620 \\
\midrule
\pairedmethod{\textbf{LMnOP (Ours)}} & 90\% & \rankfirst{0.373} & \rankfirst{0.390} & \rankfirst{0.535} & \rankfirst{0.398} & \rankfirst{0.403} & \rankfirst{0.351} \\
 & 99\% & \rankfirst{0.594} & \rankfirst{0.590} & \ranksecond{0.714} & \rankfirst{0.622} & \rankfirst{0.626} & \rankfirst{0.575} \\
\bottomrule
\end{tabularx}
\vspace{1pt}
{\raggedleft\scriptsize $^{\dagger}$Our implementation.\quad $^{*}$Importance or selection component only.\par}
\end{table}

\subsection{Pruning ratio}
\label{sec:ratio}
\begin{figure}[!t]
\centering\includegraphics[width=\linewidth]{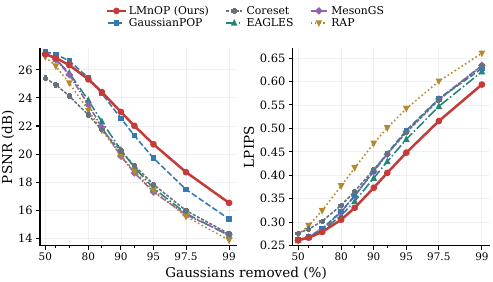}
\caption{\textbf{PSNR and LPIPS across pruning ratios.}
Mip-NeRF360, without fine-tuning.
Each point is a separate pruning of the original model.
The horizontal axis is spaced by the logarithm of the kept fraction.}
\label{fig:ratio}
\end{figure}

The comparisons above used 90\% and 99\% removal.
We also varied the pruning ratio from 50\% to 99\% on Mip-NeRF360 (\cref{fig:ratio}).
Every point is a separate pruning of the original model.
At low pruning ratios, most methods were close, and GaussianPOP had a slightly higher PSNR up to 80\% removal.
As more Gaussians were removed, the advantage of \textit{LMnOP} grew.
It had the lowest LPIPS from 70\% removal on and the highest PSNR from 85\% on.
Its PSNR margin over GaussianPOP widened from 0.45\,dB at 90\% removal to 1.15\,dB at 99\%.
This margin includes the effect of the scoring loss as well as that of the importance, and we analyze both in \cref{sec:ablation}.

\subsection{Fine-tuning after pruning}
\label{sec:finetuning}
All results above were measured without fine-tuning.
Several pruning methods, however, are evaluated after fine-tuning~\cite{lightgaussian,pup,gaussianpop}.
We therefore also fine-tuned every model pruned at 90\% removal on Mip-NeRF360 with the same 5{,}000 optimization steps on the training images.
We disabled densification and pruning, so the number of Gaussians stayed fixed.
After fine-tuning, \textit{LMnOP} was still the best on all three metrics (\cref{tab:recovery}).
Fine-tuning recovered much of the damage of pruning and narrowed the differences between the methods.
The better selection of \textit{LMnOP} nevertheless remained visible in SSIM and LPIPS.

\begin{table}[!t]
\centering\compacttable
\caption{\textbf{Fine-tuning after pruning.}
Mip-NeRF360 at 90\% removal.
All methods use the same 5{,}000 steps with a fixed number of Gaussians.
Each block is ranked separately.}
\label{tab:recovery}
\begin{tabularx}{\linewidth}{@{}l*{6}{R}}
\toprule
 & \multicolumn{3}{c}{Before fine-tuning} & \multicolumn{3}{c}{After 5,000 steps} \\
\cmidrule(lr){2-4}\cmidrule(lr){5-7}
Method & PSNR$\uparrow$ & SSIM$\uparrow$ & LPIPS$\downarrow$ & PSNR$\uparrow$ & SSIM$\uparrow$ & LPIPS$\downarrow$ \\
\midrule
EAGLES$^{*}$ & \rankthird{20.29} & \rankthird{0.651} & \ranksecond{0.394} & \rankthird{26.22} & \rankthird{0.778} & \ranksecond{0.321} \\
MesonGS$^{*}$ & 19.86 & 0.630 & \rankthird{0.408} & 25.98 & 0.769 & 0.331 \\
GaussianPOP$^{\dagger}$ & \ranksecond{22.55} & \ranksecond{0.664} & 0.409 & \ranksecond{26.97} & \ranksecond{0.786} & \rankthird{0.327} \\
RAP & 19.89 & 0.568 & 0.468 & 25.92 & 0.743 & 0.374 \\
Coreset$^{*}$ & 20.22 & 0.646 & 0.412 & 26.08 & 0.773 & 0.332 \\
\midrule
\textbf{LMnOP (Ours)} & \rankfirst{23.00} & \rankfirst{0.715} & \rankfirst{0.373} & \rankfirst{27.01} & \rankfirst{0.801} & \rankfirst{0.304} \\
\bottomrule
\end{tabularx}
\vspace{1pt}
{\raggedleft\scriptsize $^{\dagger}$Our implementation.\quad $^{*}$Importance or selection component only.\par}
\end{table}

\subsection{Ablation}
\label{sec:ablation}
Finally, we analyze the components of \textit{LMnOP} on Mip-NeRF360 without fine-tuning (\cref{tab:ablation}).
We start from leave-one-out importance under the squared error in a single pruning stage (first row).
Replacing the squared error with our scoring loss improved all three metrics at both pruning ratios (second row).
We then compared two ways to account for Gaussians that are removed together.
With leave-one-out importance, we pruned in eight stages and scored the remaining Gaussians again after each stage (third row).
\textit{LMnOP} instead uses LMO importance and prunes once after the $R$ importance computations (fourth row).

The scoring loss improved the pruning quality, and so did both ways of accounting for Gaussians that are removed together.
With the same scoring loss and a single pruning stage, LMO importance raised PSNR over leave-one-out importance by 2.23 and 2.48\,dB (second and fourth rows).
\textit{LMnOP} was also close to the eight-stage variant.
At 90\% removal, \textit{LMnOP} was better in SSIM and LPIPS and lower by only 0.08\,dB in PSNR.
At 99\% removal, the two had the same LPIPS, and \textit{LMnOP} was lower by 0.28\,dB in PSNR and 0.003 in SSIM.

\begin{table}[!t]
  \centering\compacttable
  \caption{\textbf{Ablation.}
Mip-NeRF360 without fine-tuning.
The first row is the baseline, leave-one-out importance under the squared error in a single pruning stage, and the last row is \textit{LMnOP}.}
  \label{tab:ablation}
  \begin{tabularx}{\linewidth}{@{}c@{\hspace{8pt}}c@{\hspace{8pt}}c*{6}{R}}
  \toprule
  \multicolumn{3}{c}{Components} & \multicolumn{3}{c}{90\% removed} & \multicolumn{3}{c}{99\% removed} \\\cmidrule(lr){1-3}\cmidrule(lr){4-6}\cmidrule(lr){7-9}
  \shortstack{Scoring\\loss} & \shortstack{Pruning\\8 times} & LMO & PSNR$\uparrow$ & SSIM$\uparrow$ & LPIPS$\downarrow$ & PSNR$\uparrow$ & SSIM$\uparrow$ &
  LPIPS$\downarrow$ \\
  \midrule
  \multicolumn{3}{l}{Leave-one-out, squared error} & 20.17 & 0.639 & 0.421 & 13.39 & 0.370 & 0.636 \\
  \yes & -- & -- & \rankthird{20.77} & \rankthird{0.675} & \ranksecond{0.387} & \rankthird{14.05} & \rankthird{0.405} & \rankthird{0.621} \\
  \yes & \yes & -- & \rankfirst{23.08} & \ranksecond{0.699} & \rankthird{0.389} & \rankfirst{16.81} & \rankfirst{0.470} & \ranksecond{0.594} \\
  \yes & -- & \yes & \ranksecond{23.00} & \rankfirst{0.715} & \rankfirst{0.373} & \ranksecond{16.53} & \ranksecond{0.467} & \rankfirst{0.594} \\
  \bottomrule
  \end{tabularx}
  \end{table}

\section{Limitations}
On Mip-NeRF360, \textit{LMnOP} took 19.5 and 17.0 minutes per scene at 90\% and 99\% removal, whereas GaussianPOP took 0.9 and 0.6 minutes.
Part of this gap comes from the scoring loss.
The eight-stage leave-one-out variant of \cref{tab:ablation} took 10.7 and 7.4 minutes.
\textit{LMnOP} is about twice as slow as this variant, because it keeps all the Gaussians during its $R$ importance computations.
The eight-stage variant was also slightly better than \textit{LMnOP} in PSNR and SSIM at 99\% removal.
The advantage of \textit{LMnOP} is also small at low pruning ratios and for original models that are already compact.
Combining LMO importance with several pruning stages is left for future work.

\section{Conclusion}
We presented \textit{LMnOP}, which estimates the importance of each Gaussian when the other Gaussians are removed as well.
Its $\mu$-$\sigma$ rendering makes this computation tractable by giving the exact mean and variance of each pixel's color.
Without fine-tuning, \textit{LMnOP} achieved the best results in 16 of 18 metrics across three datasets at 90\% and 99\% removal.
Our analysis further showed that the importance of a Gaussian depends on which Gaussians remain.
These results support evaluating each Gaussian in the context of the subsets that pruning is expected to produce.

\clearpage
{
    \section*{LLM Usage Statement}
    We used ChatGPT (OpenAI) and Claude (Anthropic) in this work.
    For the text, the related work, the figures, and the research ideas, the tools edited and refined content produced by the authors.
    For code and experiments, the tools generated implementation code and helped design, run, and analyze the experiments, and the authors reviewed, edited, and approved all of this work.
    We also gave the manuscript to both tools to obtain reviewer-style feedback.
    No LLM was used to create or annotate data.
    The authors verified every derivation, result, and citation and take full responsibility for the paper.
    
    \small
    \bibliographystyle{ieeenat_fullname_citeorder}
    \bibliography{main}
}
\end{document}